\documentclass[runningheads]{llncs}
\usepackage{graphicx}
\usepackage{comment}
\usepackage{color}
\usepackage{booktabs}

\usepackage{amsmath,amssymb} 
\usepackage{amsmath}
\usepackage{siunitx}

\DeclareMathOperator*{\argmax}{argmax} 
\DeclareMathOperator*{\argmin}{argmin} 

\newcommand{\partialdomain}{\mathcal{P}}
\newcommand{\completedomain}{\mathcal{C}}
\newcommand{\partialshape}{\textbf{P}}
\newcommand{\completeshape}{\textbf{C}}
\newcommand{\completionshape}{\hat{\completeshape}}

\newcommand{\codex}{\textbf{x}}
\newcommand{\partialspace}{\mathbb{X}_p}
\newcommand{\partialcode}{\codex_p}
\newcommand{\completespace}{\mathbb{X}_c}
\newcommand{\completecode}{\codex_c}
\newcommand{\completioncode}{\hat{\codex}_c}
\newcommand{\generator}{G}
\newcommand{\discriminator}{F}
\newcommand{\modez}{\textbf{z}}
\newcommand{\completecodevae}{\codex^{\text{v}}_c}
\newcommand{\completespacevae}{\mathbb{X}^{\text{v}}_c}
\newcommand{\Ez}{E_z}
\newcommand{\zrecon}{\tilde{\modez}}
\newcommand{\GaussianDistribution}{\mathcal{N}(0, \mathcal{I})}
\newcommand{\Ecomp}{\Eae}
\newcommand{\Dcomp}{\Dae}
\newcommand{\Eae}{E_{\text{AE}}}
\newcommand{\Dae}{D_{\text{AE}}}
\newcommand{\Evae}{E_{\text{VAE}}}
\newcommand{\Dvae}{D_{\text{VAE}}}

\newcommand{\LossGAN}{\mathcal{L}^{\text{GAN}}}
\newcommand{\LossRecon}{\mathcal{L}^{\text{recon}}}
\newcommand{\LossLatent}{\mathcal{L}^{\text{latent}}}
\newcommand{\LossKL}{\mathcal{L}^{\text{KL}}}

\newcommand\norm[1]{\left\lVert#1\right\rVert}

\newcommand{\Oimlz}{Ours-im-l2z}
\newcommand{\Oimpcz}{Ours-im-pc2z}

\begin{document}
\pagestyle{headings}
\mainmatter
\def\ECCVSubNumber{1915}  

\title{Multimodal Shape Completion via Conditional Generative Adversarial Networks} 

\titlerunning{Multimodal Shape Completion}
%
\author{Rundi Wu\inst{1}\thanks{Equal contribution} \and
Xuelin Chen\inst{2\star} \and
Yixin Zhuang\inst{1} \and 
Baoquan Chen\inst{1}}
%
%
\institute{Center on Frontiers of Computing Studies, Peking University \and
Shandong University\\}
\maketitle

\begin{abstract}
Several deep learning methods have been proposed for completing partial data from shape acquisition setups, i.e., filling the regions that were missing in the shape.
These methods, however, only complete the partial shape with a single output, ignoring the ambiguity when reasoning the missing geometry.
Hence, we pose a \emph{multi-modal} shape completion problem, in which we seek to complete the partial shape with multiple outputs by learning a one-to-many mapping.
We develop the first multimodal shape completion method that completes the partial shape via conditional generative modeling, without requiring paired training data.
Our approach distills the ambiguity by conditioning the completion on a learned multimodal distribution of possible results.
We extensively evaluate the approach on several datasets that contain varying forms of shape incompleteness, and compare among several baseline methods and variants of our methods qualitatively and quantitatively, demonstrating the merit of our method in completing partial shapes with both diversity and quality.

\end{abstract}

\section{Introduction}

Shape completion, which seeks to reason the geometry of the missing regions in incomplete shapes, is a fundamental problem in the field of computer vision, computer graphics and robotics. 
A variety of solutions now exist for efficient shape acquisition. The acquired shapes, however, are often incomplete, e.g., incomplete work in the user modeling interface, and incomplete scans resulted from occlusion.
The power of shape completion enables the use of these incomplete data in downstream applications, e.g., virtual walk-through, intelligent shape modeling, path planning.

\begin{figure*}[t]
   \centering
      {\includegraphics[width=0.95\linewidth]{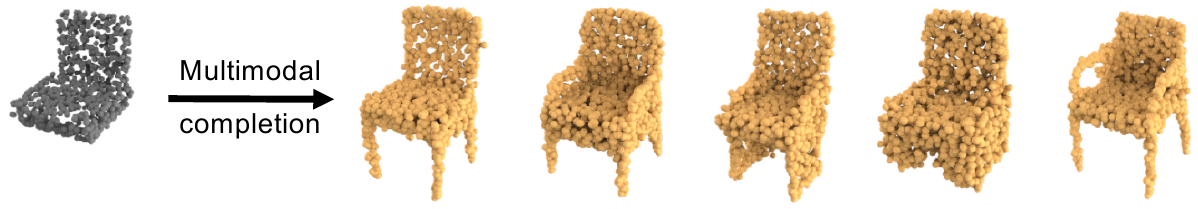} }
      \caption{We present a point-based shape completion network that can complete the partial shape with multiple plausible results. Here we show a sampling of our results, where a partial chair is completed with different types of chairs.}
   \label{fig:multimodal_completions}
\end{figure*}

With the rapid progress made in deep learning, many data-driven methods have been proposed and demonstrated effective in shape completion~\cite{chen_pcl2pcl2020,dai2017shape,gurumurthy2019high_cmu,han2017high,liu2019morphing,sharma2016vconv,stutz2018learning,thanh2016field,wang20173dshape_inpaint,wu2019pq,wu20153d,yang20183d-recgan,yu2018punet,yuan2018pcn}. 
However, most approaches in this topic have focused on completing the partial shape with a \emph{single} result, learning a one-to-one mapping for shape completion. 
In contrast, we model a distribution of potential completion results, as the shape completion problem is \emph{multimodal} in nature, especially when the incompleteness causes significant ambiguity.
For example, as shown in Fig.~\ref{fig:multimodal_completions}, a partial chair can be completed with different types of chairs.
Hence, we pose the multimodal shape completion problem, which seeks to associate each incomplete shape with multiple different complete shapes. 

In this work, we propose a first point-based multimodal shape completion method, in which the multimodality of potential outputs is distilled in a low-dimensional latent space, enabling random sampling of completion results at inference time.
The challenge to achieve this is two-fold.
First, modeling the multimodality in the high-dimensional shape space and mapping it into a low-dimensional latent space is challenging. A common problem is \emph{mode collapse}, where only a subset of the modes are represented in the low-dimensional space.
Second, the ground-truth supervision data, on which most learning-based methods rely for success, is extremely hard to acquire in our problem. Without the availability of particular supervision data (i.e., for each training incomplete shape, multiple corresponding complete shapes are required), it is challenging to learn the multimodal mapping for shape completion in an unsupervised manner \emph{without} any paired data.

We address the challenge by completing the partial shape in a conditional generative modeling setting.
We design a conditional generative adversarial network (cGAN) wherein a generator learns to map incomplete training data, combined with a latent vector sampled from a learned multimodal shape distribution, to a suitable latent representation such that a discriminator cannot differentiate between the mapped latent variables and the latent variables obtained from complete training data (i.e., complete shape models).
An encoder is introduced to encode mode latent vectors from complete shapes, learning the multimodal distribution of all possible outputs.
We further apply this encoder to the completion output to extract and recover the input latent vector, forcing the bijective mapping between the latent space and the output modes.
The mode encoder is trained to encode the multimodality in an explicit manner (Section~\ref{sec:exp_encoder}), alleviating the aforementioned mode collapse issue.
By conditioning the generation of completion results on the learned multimodal shape distribution, we achieve multimodal shape completion.

We extensively evaluate our method on several datasets that contain varying forms of shape incompleteness.
We compare our method against several baseline methods and variants of our method, and rate the different competing methods using a combination of established metrics.
Our experiments demonstrate the superiority of our method compared
to other alternatives, producing completion results with both high diversity and quality, all the while remaining faithful to the partial input shape.

\section{Related work}

\paragraph{Shape completion}
With the advancement of deep learning in 3D domain, many deep learning methods have been proposed to address the shape completion challenge. 
Following the success of CNN-based 2D image completion networks, 3D convolutional neural networks applied on voxelized inputs have been widely adopted for 3D shape completion task~\cite{dai2017shape,han2017high,sharma2016vconv,stutz2018learning,thanh2016field,wang20173dshape_inpaint,wu2019pq,wu20153d,yang20183d-recgan}. 
To avoid geometric information loss resulted from quantizing shapes into voxel grids, several approaches~\cite{chen_pcl2pcl2020,gurumurthy2019high_cmu,liu2019morphing,yu2018punet,yuan2018pcn} have been develop to work directly on point sets to reason the missing geometry.
While all these methods resort to learning a parameterized model (i.e., neural networks) as a mapping from incomplete shapes to completed shapes, the learned mapping function remains injective. 
Consequently, these methods can only complete the partial shape with a single deterministic result, ignoring the ambiguity of the missing regions.

\paragraph{Generative modeling}
The core of generative modeling is parametric modeling of the data distribution. 
Several classical approaches exist for tackling the generative modeling problem, e.g., restricted Boltzmann machines~\cite{smolensky1986information}, and autoencoders~~\cite{hinton2006reducing,vincent2008extracting}. 
Since the introduction of Generative Adversarial Networks (GANs)~\cite{GAN_NIPS2014_5423}, it has been widely adopted for a variety of generative tasks. 
In 2D image domain, researchers have utilized GANs in tasks ranging from image generation~\cite{lsgan,dcgan,wgan}, image super-resolution~\cite{ledig2017photo}, to image inpainting in 2D domain.
In the context of 3D, a lot of effort has also been put on the task of content generation~\cite{achlioptas2018learning,im-net,atlasnet,park2019deepsdf}. 
The idea of using GANs to generatively model the missing regions for shape completion has also been explored in the pioneering works~\cite{chen_pcl2pcl2020,gurumurthy2019high_cmu}. In this work, we achieve multimodal shape completion by utilizing GANs to reason the missing geometry in a conditional generative modeling setting.

\paragraph{Deep learning on points}
Our method is built upon recent success in deep neural networks for point cloud representation learning.
Although many improvements to PointNet~\cite{qi2017pointnet} have been proposed~\cite{li2018so,li2018pointcnn,qi2017pointnet++,su2018splatnet,zaheer2017deep}, the simplicity and effectiveness of PointNet and its extension PointNet++ make them popular in many other analysis tasks~\cite{guerrero2018pcpnet,yin2018p2p,yu2018ec,yu2018punet}. 
To achieve point cloud generation,~\cite{achlioptas2018learning} proposed to train GANs in the latent space produced by a PointNet-based autoencoder, reporting significant performance gain in point cloud generation.
Similar to~\cite{chen_pcl2pcl2020}, we also leverage the power of such point cloud generation model to complete shapes, but conditioning the generation on a learned distribution of potential completion outputs for multimodal shape completion.

\section{Method}
\begin{figure*}[t]
    \centering
    \includegraphics[width=0.95\textwidth]{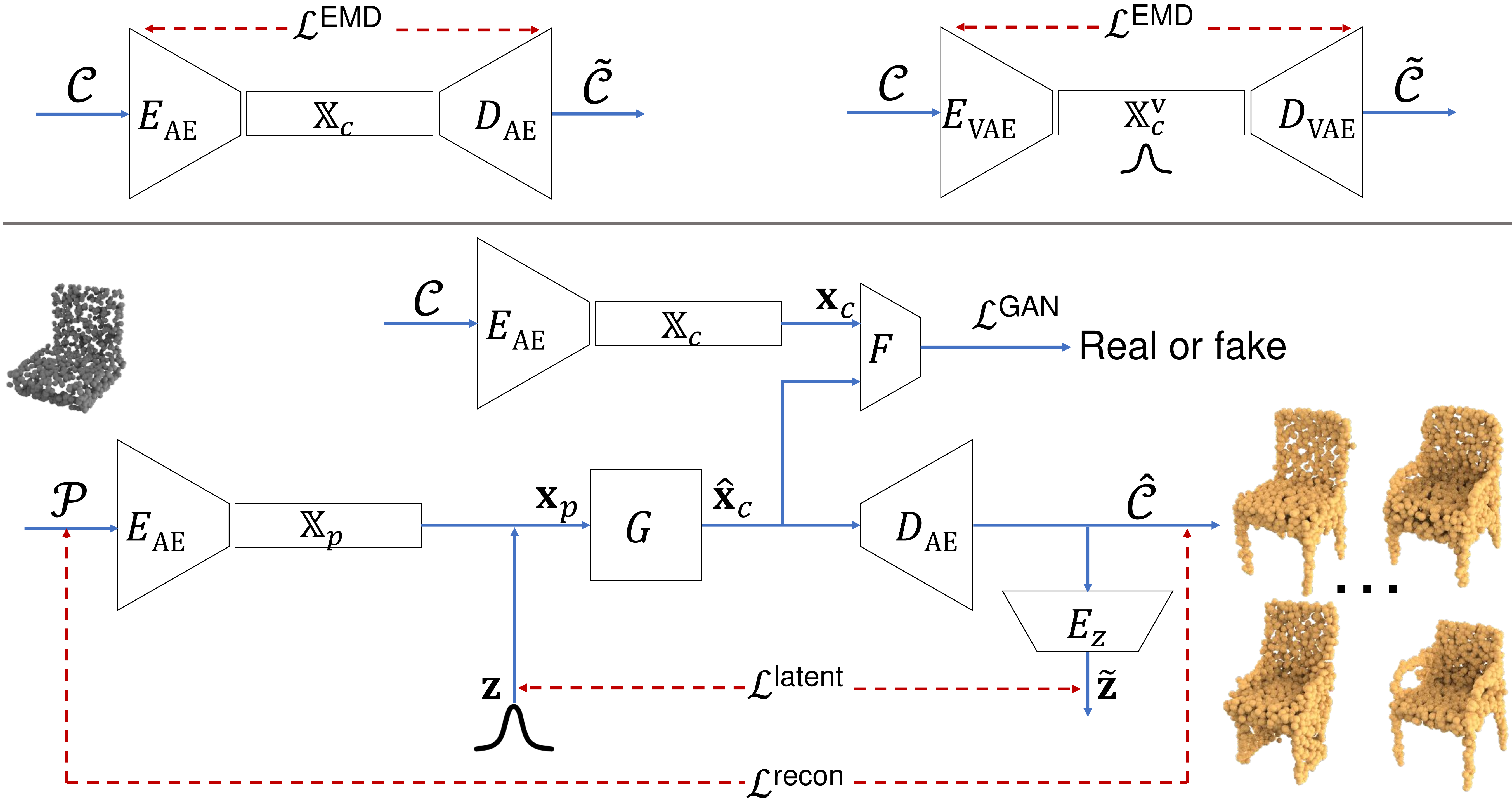}
    \caption{The proposed network architecture for learning multimodal shape completion. We use the encoder $\Evae$ of a shape variational autoencoder as the mode encoder $\Ez$ to encode the multimodality from shapes explicitly.}
    \label{fig:network}
\end{figure*}
Given a partial shape domain $\partialdomain \subset \mathbb{R}^{K \times 3}$, we seek to learn a multimodal mapping from $\partialdomain$ to the complete shape domain $\completedomain \subset \mathbb{R}^{N \times 3}$, achieving the goal of multimodal shape completion.
Unlike existing methods, which are primarily limited to producing a single deterministic completion output $\completionshape$ from a partial shape $\partialshape \in \partialdomain$, our method learns a mapping that could sample the completion $\completionshape$ from the conditional distribution of possible completions, producing diverse completion results. 
During training, our method only requires access to a set of partial point sets, and a set of complete point sets.
It is important to note that there is no any paired completion instances in $\completedomain$ for point sets in $\partialdomain$.

Following the spirit of~\cite{chen_pcl2pcl2020}, without paired training data, we address the multimodal shape completion problem via adversarial training on the latent space learned from point sets, while introducing a low-dimensional code $\modez \in \mathbb{R}^{\text{Z}}$, of which the latent space is learned by modeling the multimodality presented in possible outputs, as a conditional input in addition to the partial shape. To enable stochastic sampling, we desire $\modez$ to be drawn from a prior distribution $p(\modez)$; a standard Gaussian distribution $\GaussianDistribution$ is used in this work.

More specifically, we learn two class-specific point set manifolds, $\partialspace$ for the partial shapes, and $\completespace$ for the complete shapes. 
Solving the multimodal shape completion problem then amounts to learning a mapping $\partialspace \to \completespace$ in a conditional generative modeling setting between respective latent spaces. 
Hence, we train a generator $\generator \colon (\partialspace, p(\modez)) \to \completespace$, to perform the multimodal mapping with the latent code $\modez$ as a condition input. 
In absence of paired training data, we opt to use adversarial training to complete shapes in a generative modeling setting.
Furthermore, to force the generator to use the latent code $\modez$, we introduce a encoder $\Ez$ to recover $\modez$ from the completion output, forcing the connection between the latent space and the shape space to be invertible.
Fig.~\ref{fig:network} shows the setup of the proposed multimodal shape completion network. All network modules are detailed next.

\subsection{Learning latent spaces for point sets}

The latent space of a given set of point sets is obtained by training an autoencoder, which encodes the given input to a low-dimension latent feature and then decodes to reconstruct the original input.

For point sets coming from the complete point sets $\completedomain$, we learn an encoder network $\Ecomp$ that maps $\completeshape$ from the original parameter space $\mathbb{R}^{N \times 3}$, defined by concatenating the coordinates of the $N$ points, to a lower-dimensional latent space $\completespace$. A decoder network $\Dcomp$ performs the inverse transformation back to $\mathbb{R}^{N \times 3}$ giving us a reconstructed point set $\tilde{\completeshape}$ with also $N$ points. The encoder-decoders are trained with reconstruction loss:
\begin{equation}
\mathcal{L}^{\textrm{EMD}} = \mathbb{E}_{\completeshape \sim p(\completeshape)} d^{\text{EMD}}( \completeshape,  \Dcomp (\Ecomp (\completeshape)) ), 
\end{equation}
where $\completeshape \sim p(\completeshape)$ denotes point set samples drawn from the set of complete point sets, $d^{\text{EMD}}(X_1, X_2)$ is the Earth Mover's Distance~(EMD) between point sets $X_1, X_2$.  Once trained, the network weights are held fixed and the \emph{complete} latent code $\completecode = \Ecomp(\completeshape),\ \completecode \in \completespace$ for a complete point set $\completeshape$ provides a compact representation for subsequent training and implicitly captures the manifold of complete data.
As for the point set coming from the partial point sets $\partialdomain$, instead of training another autoencoder for its latent parameterization, we directly feed the partial point sets to $\Ecomp$ obtained above for producing \emph{partial} latent space $\partialspace$, which in~\cite{chen_pcl2pcl2020} is proved to yield better performance in subsequent adversarial training. Note that, to obtain $\partialcode \in \partialspace$, we duplicate the partial point set of $K$ points to align with the number of complete point set before feed it to $\Eae$.

\subsection{Learning multimodal mapping for shape completion}

Next, we setup a min-max game between the generator and the discriminator to perform the multimodal mapping between the latent spaces. 
The generator is trained to fool the discriminator such that the discriminator fails to reliably tell if the latent variable comes from original $\completespace$ or the remapped $\partialspace$. 
The mode encoder is trained to model the multimodal distribution of possible complete point sets, and is further applied to the completion output to encode and recover the input latent vector, encouraging the use of conditional mode input. 

Formally, the latent representation of the input partial shape $\partialcode = \Ecomp(\partialshape)$, along with a Gaussian-sampled condition $\modez$, is mapped by the generator to $\completioncode = \generator(\partialcode, \modez)$. Then, the task of the discriminator $\discriminator$ is to distinguish between latent representations $\completioncode$ and $\completecode = \Ecomp(\completeshape)$.
The mode encoder $\Ez$ will encode the completion point set, which can be decoded from the completion latent code $\completionshape = \Dcomp(\completioncode)$, to reconstruct the conditional input $\zrecon = \Ez(\completionshape)$.
We train the mapping function using a GAN. Given training examples of complete latent variables $\completecode$, remapped partial latent variables $\completioncode$, and Gaussian samples $\modez$, we seek to optimize the following training losses over the generator $\generator$, the discriminator $\discriminator$, and the encoder $\Ez$:

\textbf{Adversarial loss.} We add the adversarial loss to train the generator and discriminator. In our implementation, we use least square GAN~\cite{lsgan} for stabilizing the training. Hence, the adversarial losses minimized for the generator and the discriminator are defined as:
\begin{align}
    \LossGAN_\discriminator & = \mathbb{E}_{\completeshape \sim p(\completeshape)}
    {\left[{\discriminator 
    {\left(  \Ecomp(\completeshape) \right)}} -1 \right]^2}  
     + \mathbb{E}_{\partialshape \sim p(\partialshape), \modez \sim p(\modez) } {\left[ {\discriminator{\left(
   \generator( \Ecomp ( \partialshape ), \modez ) \right)}} 
   \right]^2}   \\
   \LossGAN_\generator & = 
   \mathbb{E}_{\partialshape \sim p(\partialshape), \modez \sim p(\modez) } {\left[
   \discriminator{\left(
   \generator( \Ecomp ( \partialshape ), \modez ) \right)} -1
   \right]^2},
   \label{eqn:lsgan_loss}
\end{align}
where $\completeshape \sim p(\completeshape)$, $\partialshape \sim p(\partialshape)$ and $\modez \sim p(\modez)$ denotes samples drawn from, respectively, the set of complete point sets, the set of partial point sets, and $\GaussianDistribution$.

\textbf{Partial reconstruction loss.}
Similar to previous work~\cite{chen_pcl2pcl2020}, we add a reconstruction loss to encourage the generator to \emph{partially} reconstruct the partial input, so that the completion output is faithful to the input partial:
\begin{equation}
    \LossRecon_\generator = \mathbb{E}_{\partialshape \sim p(\partialshape), \modez \sim p(\modez) } {\left[ d^{\text{HL}}(\partialshape, \Dcomp(\generator(\Ecomp(\partialshape), \modez)) \right]} ),
    \label{eqn:partial_recon_loss}
\end{equation}
where $d^{\text{HL}}$ denotes the unidirectional Hausdorff distance from the partial point set to the completion point set.

\textbf{Latent space reconstruction.}
A reconstruction loss on the $\modez$ latent space is also added to encourage $G$ to use the conditional mode vector $\modez$:
\begin{equation}
    \LossLatent_{{\generator, \Ez}} = \mathbb{E}_{\partialshape \sim p(\partialshape), \modez \sim p(\modez) } {\left[ \norm{\modez, \Ez(\Dcomp(\generator(\Ecomp(\partialshape), \modez))) }_1 \right]},
    \label{eqn:partial_recon_loss}
\end{equation}

Hence, our full objective function for training the multimodal shape completion network is described as:
\begin{equation}
    \argmin_{(\generator, \Ez)} \argmax_{\discriminator} \LossGAN_\discriminator + \LossGAN_\generator + \alpha \LossRecon_\generator + \beta \LossLatent_{{\generator, \Ez}},
    \label{eqn:obj_function}
\end{equation}
where $\alpha$ and $\beta$ are importance weights for the partial reconstruction loss and the latent space reconstruction loss, respectively.

\subsection{Explicitly-encoded multimodality}
\label{sec:exp_encoder}
To model the multimodality of possible completion outputs, we resort to an explicit multimodality encoding strategy, 
in which $\Ez$ is trained as a part to explicitly reconstruct the complete shapes.
More precisely, a variational autoencoder $(\Evae, \Dvae)$ is pre-trained to encode complete shapes into a standard Gaussian distribution $\GaussianDistribution$ and then decode to reconstruct the original shapes.
Once trained, $\Evae$ can encode complete point sets as $\completecodevae \in \completespacevae$, and can be used to recover the conditional input $\modez$ from the completion output.
Hence, the mode encoder is set to $\Ez = \Evae$ and held fixed during the GAN training.

Another strategy is to implicitly encode the multimodality, in which the $\Ez$ is jointly trained to map complete shapes into a latent space without being trained as a part of explicit reconstruction. Although it has been shown effective to improve the diversity in the multimodal mapping learning~\cite{zhu2017toward,von}, 
we demonstrate that using explicitly-encoded multimodality in our problem yields better performance.
In Section~\ref{sec:comparison}, we present the comparison against variants of using implicit multimodality encoding.

\subsection{Implementation details}
In our experiments, a partial shape is represented by $K=1024$ points and a complete shape by $N=2048$ points.
The the point set (variational) autoencoder follows \cite{achlioptas2018learning,chen_pcl2pcl2020}: using a PointNet\cite{qi2017pointnet} as the encoder and a 3-layer MLP as the decoder. 
The autoencoder encodes a point set into a latent vector of fixed dimension $|\textbf{x}|=128$.
Similar to \cite{achlioptas2018learning,chen_pcl2pcl2020}, we use a 3-layer MLP for both generator $\generator$ and discriminator $\discriminator$.
The $\Ez$ also uses the PointNet to map a point set into a latent vector $\modez$, of which the length we set to $|\modez|=64$.
Unless specified, the trade-off parameters $\alpha$ and $\beta$ in Eq.~\ref{eqn:obj_function} are set to 6 and 7.5, respectively, in our experiments. 
For training the point set (variational) autoencoder, we use the Adam optimizer\cite{kingma2014adam} with an initial learning rate $0.0005$, $\beta_1=0.9$ and train 2000 epochs with a batch size of 200. To train the GAN, we use the Adam optimizer with an initial learning rate $0.0005$, $\beta_1=0.5$ and train for a maximum of 1000 epochs with a batch size of 50.
More details about each network module are presented in the supplementary material.

\section{Experiments}

\begin{figure*}[t]
    \centering
    \includegraphics[width=0.95\textwidth]{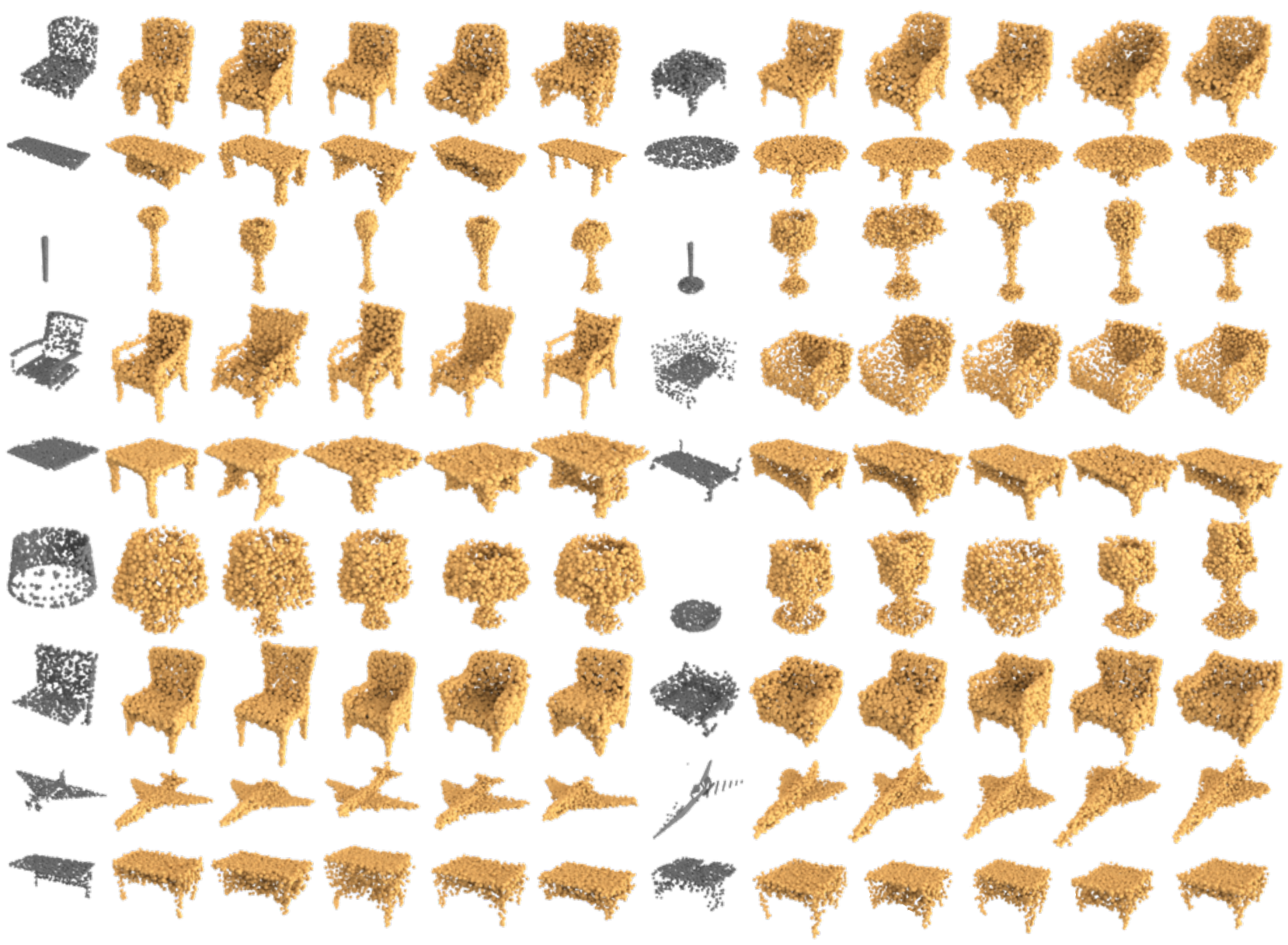}
    \caption{Our multimodal shape completion results. We show result examples, where the input partial shape is colored in grey and is followed by five different completions in yellow. From top to bottom: \texttt{PartNet} (rows 1-3), \texttt{PartNet-Scan} (rows 4-6), and \texttt{3D-EPN} (rows 7-9).}
    \label{fig:results-gallery}
\end{figure*}

\begin{figure*}[t]
    \centering
    \includegraphics[width=0.9\textwidth]{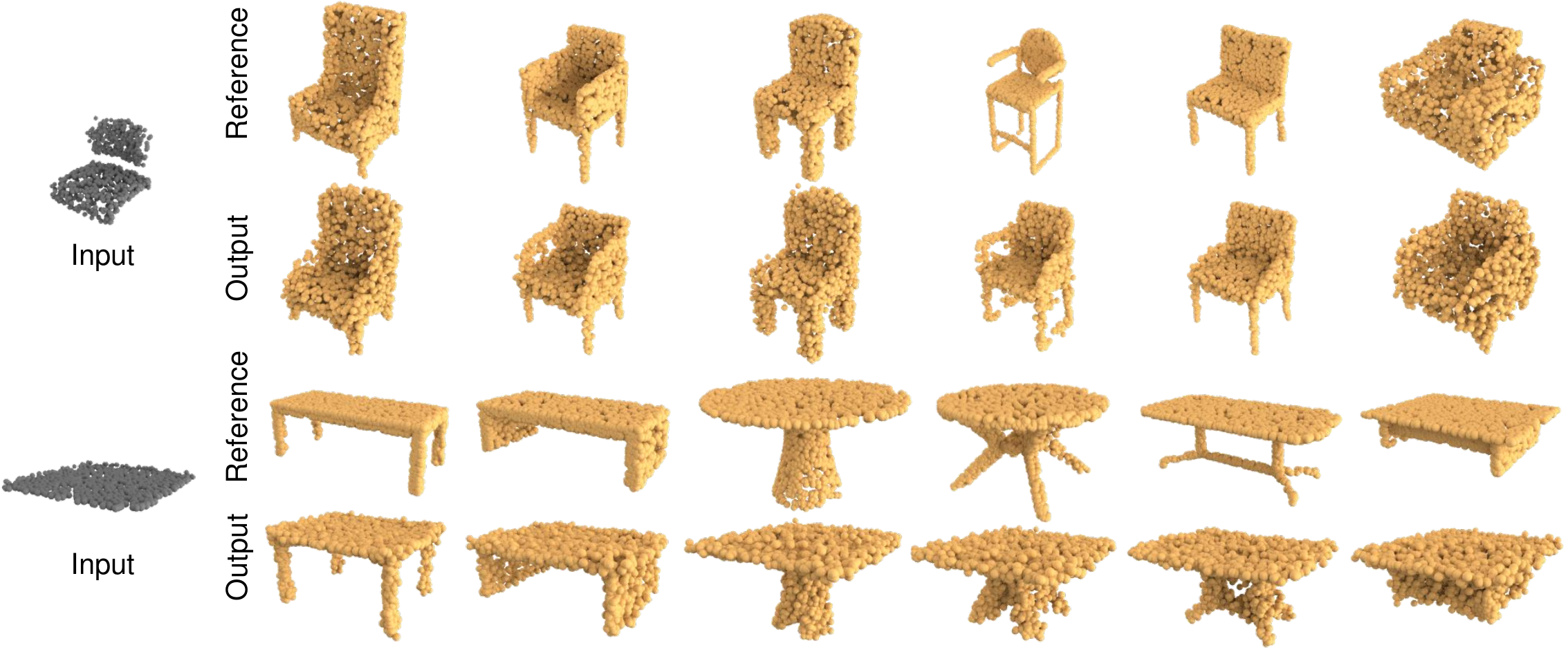} 
    \caption{Shape completion guided by reference shapes. The completion result varies accordingly when the reference shape changes.}
    \label{fig:complete-by-reference}
\end{figure*}

In this section, we present results produced from our method on multimodal shape completion, and both quantitative and qualitative comparisons against several baseline methods and variants of our method, along with a set of experiments for evaluating different aspects of our method.

\textbf{Datasets.}
Three datasets are derived to evaluate our method under different forms of shape incompleteness:
(A) \texttt{PartNet} dataset simulates part-level incompleteness in the user modeling interface. With the semantic segmentation provided in the original PartNet dataset~\cite{Mo_2019_CVPR}, for each provided point set, we remove points of randomly selected parts to create a partial point set with at least one part.
(B) \texttt{PartNet-Scan} dataset resembles the scenario where the partial scan suffers from part-level incompleteness. For each shape in~\cite{Mo_2019_CVPR}, we randomly remove parts and virtually scan residual parts to obtain a partial scan with part-level incompleteness.
(C) \texttt{3D-EPN} dataset~\cite{dai2017shape} is derived from ShapeNet~\cite{shapenet2015} and provides simulated partial scans with arbitrary incompleteness. Scans are represented as Signed Distance Field but we only use the provided point cloud representations.
Last, the complete point sets provided in PartNet~\cite{Mo_2019_CVPR} serve as the complete training data for the first two datasets, while for \texttt{3D-EPN} dataset, we use the complete virtual scan of ShapeNet objects as the complete training data.
We use Chair, Table and Lamp categories for \texttt{PartNet} and \texttt{PartNet-scan}, and use Chair, Airplane and Table categories for \texttt{3D-EPN}. In all our experiments, we train separate networks for each category in each dataset. More details about data processing can be found in the supplementary material.

\begin{table}[t]
    \centering
    \resizebox{0.95\textwidth}{!}{
        \begin{tabular}{l|cccc|cccc|cccc}
            \toprule
            \texttt{PartNet} & \multicolumn{4}{c|}{\texttt{MMD} (lower is better)} & \multicolumn{4}{c|}{\texttt{TMD} (higher is better)} & \multicolumn{4}{c}{\texttt{UHD} (lower is better)} \\
            \midrule
            Method & \multicolumn{1}{c|}{Chair} & \multicolumn{1}{c|}{Lamp} & \multicolumn{1}{c|}{Table} & Avg.  & \multicolumn{1}{c|}{Chair} & \multicolumn{1}{c|}{Lamp} & \multicolumn{1}{c|}{Table} & Avg.  & \multicolumn{1}{c|}{Chair} & \multicolumn{1}{c|}{Lamp} & \multicolumn{1}{c|}{Table} & Avg. \\
            \midrule
            \texttt{pcl2pcl} & 1.90  & 2.50  & 1.90  & 2.10  & 0.00  & 0.00  & 0.00  & 0.00  & 4.88  & 4.64  & 4.78  & \textbf{4.77} \\
            \texttt{KNN-latent} & 1.39  & 1.72  & 1.30  & \textbf{1.47} & 2.28  & 4.18  & 2.36  & 2.94  & 8.58  & 8.47  & 7.61  & 8.22 \\
            \texttt{\Oimlz} & 1.74  & 2.36  & 1.68  & 1.93  & 3.74  & 2.68  & 3.59  & \textbf{3.34} & 8.41  & 6.37  & 7.21  & 7.33 \\
            \texttt{\Oimpcz} & 1.90  & 2.55  & 1.54  & 2.00  & 1.01  & 0.56  & 0.51  & 0.69  & 6.65  & 5.40  & 5.38  & \textbf{5.81} \\
            \texttt{Ours}  & 1.52  & 1.97  & 1.46  & \textbf{1.65} & 2.75  & 3.31  & 3.30  & \textbf{3.12} & 6.89  & 5.72  & 5.56  & 6.06 \\
            \bottomrule
        \end{tabular}%
    }
    
    
    \centering
    \resizebox{0.95\textwidth}{!}{    
        \begin{tabular}{l|cccc|cccc|cccc}
            \toprule
            \texttt{PartNet-Scan} & \multicolumn{4}{c|}{\texttt{MMD} (lower is better)} & \multicolumn{4}{c|}{\texttt{TMD} (higher is better)} & \multicolumn{4}{c}{\texttt{UHD} (lower is better)} \\
            \midrule
            Method & \multicolumn{1}{c|}{Chair} & \multicolumn{1}{c|}{Lamp} & \multicolumn{1}{c|}{Table} & Avg.  & \multicolumn{1}{c|}{Chair} & \multicolumn{1}{c|}{Lamp} & \multicolumn{1}{c|}{Table} & Avg.  & \multicolumn{1}{c|}{Chair} & \multicolumn{1}{c|}{Lamp} & \multicolumn{1}{c|}{Table} & Avg. \\
            \midrule
            \texttt{pcl2pcl} & 1.96  & 2.36  & 2.09  & 2.14  & 0.00  & 0.00  & 0.00  & 0.00  & 5.20  & 5.34  & 4.73  & \textbf{5.09} \\
            \texttt{KNN-latent} & 1.40  & 1.80  & 1.39  & \textbf{1.53} & 3.09  & 4.47  & 2.85  & 3.47  & 8.79  & 8.41  & 7.50  & 8.23 \\
            \texttt{\Oimlz} & 1.79  & 2.58  & 1.92  & 2.10  & 3.85  & 3.18  & 4.75  & \textbf{3.93} & 7.88  & 6.39  & 7.40  & 7.22 \\
            \texttt{\Oimpcz} & 1.65  & 2.75  & 1.84  & 2.08  & 1.91  & 0.50  & 1.86  & 1.42  & 7.50  & 5.36  & 5.68  & \textbf{6.18} \\
            \texttt{Ours}  & 1.53  & 2.15  & 1.58  & \textbf{1.75} & 2.91  & 4.16  & 3.88  & \textbf{3.65} & 6.93  & 5.74  & 6.24  & 6.30 \\
            \bottomrule
        \end{tabular}
    }
    
    
    \centering
    \resizebox{0.95\textwidth}{!}{
        \begin{tabular}{l|cccc|cccc|cccc}
            \toprule
            \texttt{3D-EPN} & \multicolumn{4}{c|}{\texttt{MMD} (lower is better)} & \multicolumn{4}{c|}{\texttt{TMD} (higher is better)} & \multicolumn{4}{c}{\texttt{UHD} (lower is better)} \\
            \midrule
            Method & \multicolumn{1}{c|}{Chair} & \multicolumn{1}{c|}{Plane} & \multicolumn{1}{c|}{Table} & Avg.  & \multicolumn{1}{c|}{Chair} & \multicolumn{1}{c|}{Plane} & \multicolumn{1}{c|}{Table} & Avg.  & \multicolumn{1}{c|}{Chair} & \multicolumn{1}{c|}{Plane} & \multicolumn{1}{c|}{Table} & Avg. \\
            \midrule
            \texttt{pcl2pcl} & 1.81  & 1.01  & 3.12  & 1.98  & 0.00  & 0.00  & 0.00  & 0.00  & 5.31  & 9.71  & 9.03  & \textbf{8.02} \\
            \texttt{KNN-latent} & 1.45  & 0.93  & 2.25  & \textbf{1.54} & 2.24  & 1.13  & 3.25  & 2.21  & 8.94  & 9.54  & 12.70 & 10.40 \\
            \texttt{\Oimlz} & 1.91  & 0.86  & 2.78  & 1.80  & 3.84  & 2.17  & 4.27  & \textbf{3.43} & 9.53  & 10.60 & 9.36  & 9.83 \\
            \texttt{\Oimpcz} & 1.61  & 0.91  & 3.19  & 1.90  & 1.51  & 0.82  & 1.67  & 1.33  & 8.18  & 9.55  & 8.50  & \textbf{8.74} \\
            \texttt{Ours}  & 1.61  & 0.82  & 2.57  & \textbf{1.67} & 2.56  & 2.03  & 4.49  & \textbf{3.03} & 8.33  & 9.59  & 9.03  & 8.98 \\
            \bottomrule
        \end{tabular}%
    }
    
    \vspace{5pt}
    
    \caption{Quantitative comparison results on \texttt{PartNet} (top), \texttt{PartNet-Scan} (middle) and \texttt{3D-EPN} (bottom). Top two methods on each measure are highlighted. Note that \texttt{MMD} (quality), \texttt{TMD} (diversity) and \texttt{UHD} (fidelity) presented in the tables are multiplied by $10^3$, $10^2$ and $10^2$, respectively.}
    \label{table:quan-comparison}
\end{table}

\textbf{Evaluation measures.} For each partial shape in the test set, we generate $k=10$ completion results and adopt the following measures for quantitative evaluation:
\begin{itemize}
    \item Minimal Matching Distance (\texttt{MMD}) measures the \emph{quality} of the completed shape. We calculates the Minimal Matching Distance (as described in~\cite{achlioptas2018learning}) between the set of completion shapes and the set of test shapes.
    \item Total Mutual Difference (\texttt{TMD}) measures the completion \emph{diversity} for a partial input shape, by summing up all the difference among the $k$ completion shapes of the same partial input. For each shape $i$ in the $k$ generated shapes, we calculate its average Chamfer distance $d_i^{\text{CD}}$ to the other $k - 1$ shapes. The diversity is then calculated as the sum $\sum_{i=1}^{k} d_i^{\text{CD}}$. 
    \item Unidirectional Hausdorff Distance (\texttt{UHD}) measures the completion \emph{fidelity} to the input partial. We calculate the average Hausdorff distance from the input partial shape to each of the $k$ completion results.
\end{itemize}
More in-depth description can be found in the supplementary material.

\subsection{Multimodal completion results}

We first present qualitative results of our method on multimodal shape completion, by using randomly sampled $\modez$ from the standard Gaussian distribution. 
Figure \ref{fig:results-gallery} shows a collection of our multimodal shape completion results on the aforementioned datasets. More visual examples can be found in supplementary material.

To allow more explicit control over the modes in completion results, the mode condition $\modez$ can also be encoded from a user-specified shape.
As shown in Figure~\ref{fig:complete-by-reference}, this enables us to complete the partial shape under the guidance of a given reference shape.
The quantitative evaluation of our results, along with the comparison results, is presented next.

\begin{figure*}[t]
	\centering
	\includegraphics[width=0.95\textwidth]{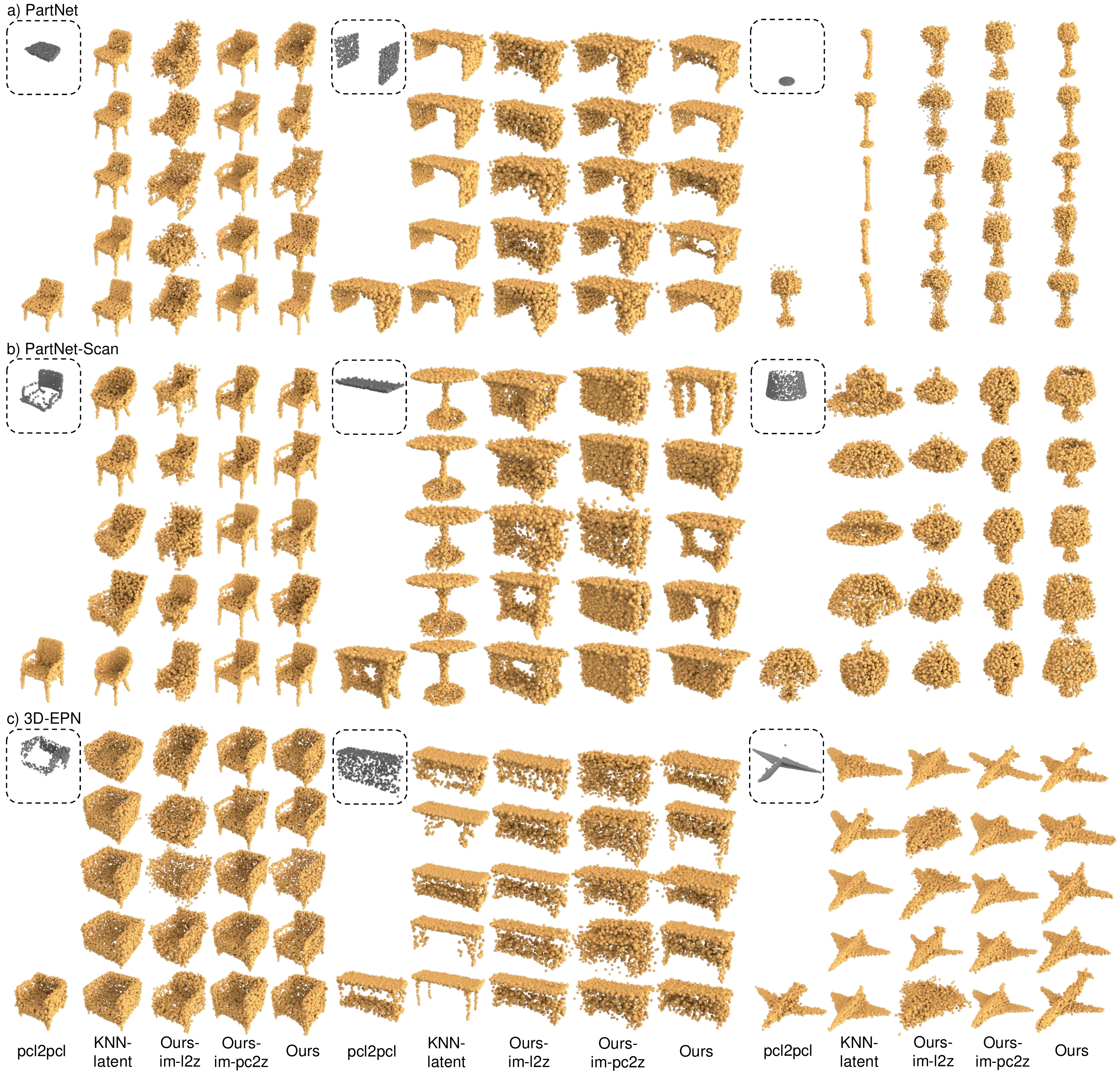}
	\caption{Qualitative comparison on each three categories of \texttt{PartNet} (top), \texttt{PartNet-Scan} (middle) and \texttt{3D-EPN} (bottom) dataset. Our method produces results that are both diverse and plausible.}
    \label{fig:visual-comparison}
\end{figure*}

\begin{figure*}[t]
    \centering
    \includegraphics[width=0.95\textwidth]{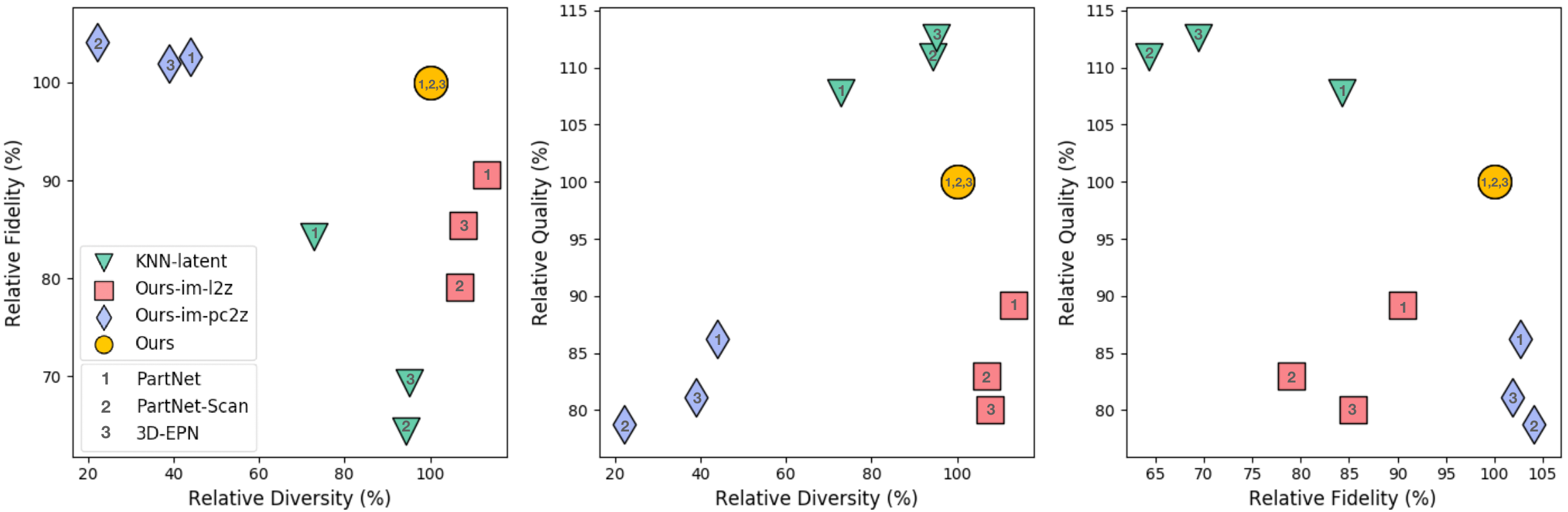}
    \caption{Comparisons using metrics combinations. 
    Our results present high diversity, quality and fidelity in comparisons using combinations of metrics.
    \emph{Relative} performance is plotted, and \texttt{pcl2plc} is excluded as it fails to present completion diversity.
    }
    \label{fig:stat-plot}
\end{figure*}

\subsection{Comparison results}
\label{sec:comparison}
We present both qualitative and quantitative comparisons against baseline methods and variants of our method:
\begin{itemize}
    \item \texttt{pcl2pcl}~\cite{chen_pcl2pcl2020}, which also uses GANs to complete via generative modeling without paired training data. Without the conditional input, this method, however, cannot complete with diverse shapes for a single partial input.
    \item \texttt{KNN-latent}, which retrieves a desired number of best candidates from the latent space formed by our complete point set autoencoder, using k-nearest neighbor search algorithm.
    \item \texttt{\Oimlz}, a variant of our method as described in Section~\ref{sec:exp_encoder}, jointly trains the $\Ez$ to implicitly model the multimodality by mapping complete data into a low-dimensional space. The $\Ez$ can either take input as complete latent codes (denoted by \texttt{\Oimlz}) or complete point clouds (denoted by \texttt{\Oimpcz}) to map to $\modez$ space. 
    \item \texttt{\Oimpcz}, in which, as stated above, the $\Ez$ takes complete point clouds as input to encode the multimodality in an implicit manner.
\end{itemize}
More details of the above methods can be found in the supplementary material.

We present quantitative comparison results in Table~\ref{table:quan-comparison}.
We can see that, by rating these methods using the combination of the established metrics, our method outperforms other alternatives with high completion quality (low \texttt{MMD}) and diversity (high \texttt{TMD}), while remains faithful to the partial input.
More specifically, 
\texttt{pcl2pcl} has the best fidelity (lowest \texttt{UHD}) to the input partial shape, the completion results, however, present no diversity;
\texttt{\Oimlz} presents the best completion diversity, but fails to produce high partial matching between the partial input and the completion results;
\texttt{\Oimpcz} suffers from severe mode collapse;
our method, by adopting the explicit multimodality encoding, can complete the partial shape with high completion fidelity, quality and diversity.

To better understand the position of our method among those competing methods when rating with the established metrics, we present Fig.~\ref{fig:stat-plot} to visualize the performance of each method in comparisons using combinations of the metrics. Note that the percentages in Fig.~\ref{fig:stat-plot} are obtained by taking the performance of our method as reference and calculating the relative performance of other methods to ours.

We also show qualitative comparison results in Fig.~\ref{fig:visual-comparison}. 
Compared to other methods, our method shows the superiority on multimodal shape completion: high completion diversity and quality while still remains faithful to the input, as consistently exhibited in the quantitative comparisons.
Note that \texttt{pcl2pcl} cannot complete the partial shape with multiple outputs, thus only a single completion result is shown in the figure.

\subsection{Results on real scans}
\begin{figure*}[h]
    \centering
    \includegraphics[width=0.9\textwidth]{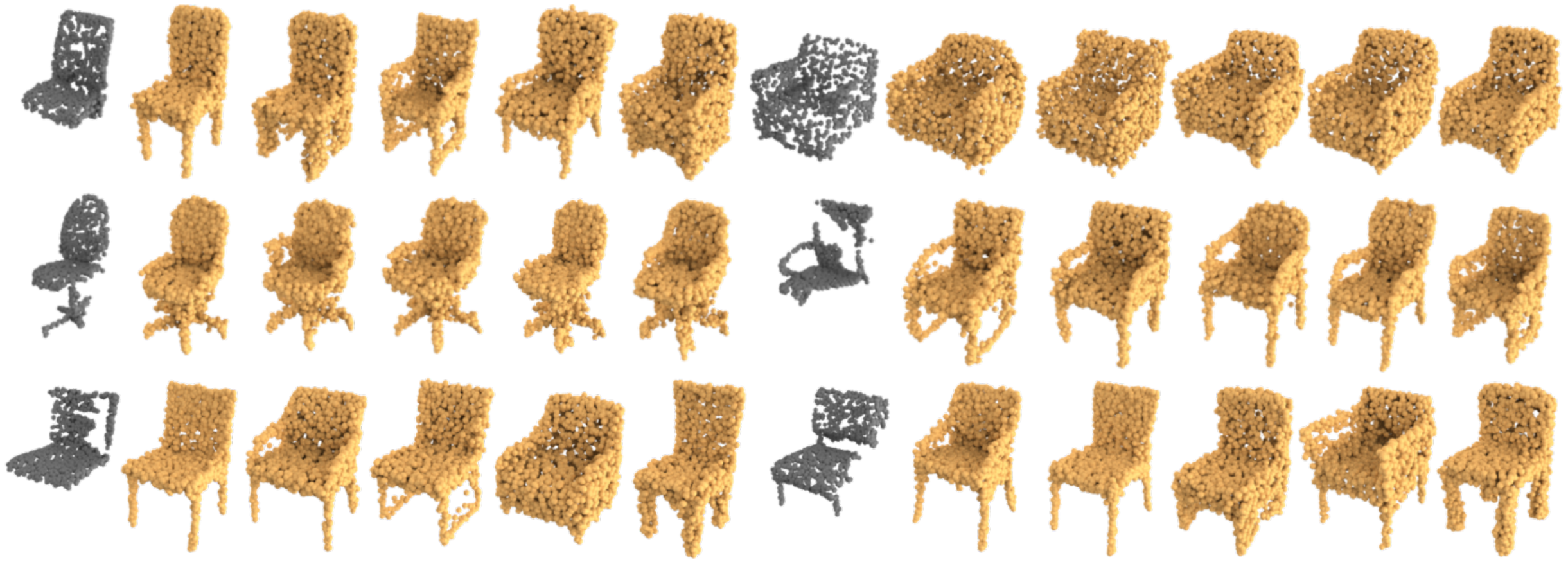}
    \caption{Examples of our multimodal shape completion results on real scan data.}
    \label{fig:real-scan}
\end{figure*}

The nature of unpaired training data setting enables our method to be directly trained on real scan data with ease. We have trained and tested our method on the real-world scans provided in \texttt{pcl2pcl}, which contains around $550$ chairs. We randomly picked $50$ of them for testing and used the rest for training.
Qualitative results are shown in Fig.~\ref{fig:real-scan}. For quantitative evaluation, our method achieved \texttt{MMD} of $2.42 \times 10^{-3}$, \texttt{TMD} of $3.17\times 10^{-2}$ and \texttt{UHD} of $8.60\times 10^{-2}$.

\subsection{More experiments}
We also present more experiments conducted to evaluate different aspects of our method.

\begin{figure*}[t]
    \centering
    \begin{minipage}[t]{0.45\textwidth}
		\centering
    	\includegraphics[width=0.9\textwidth]{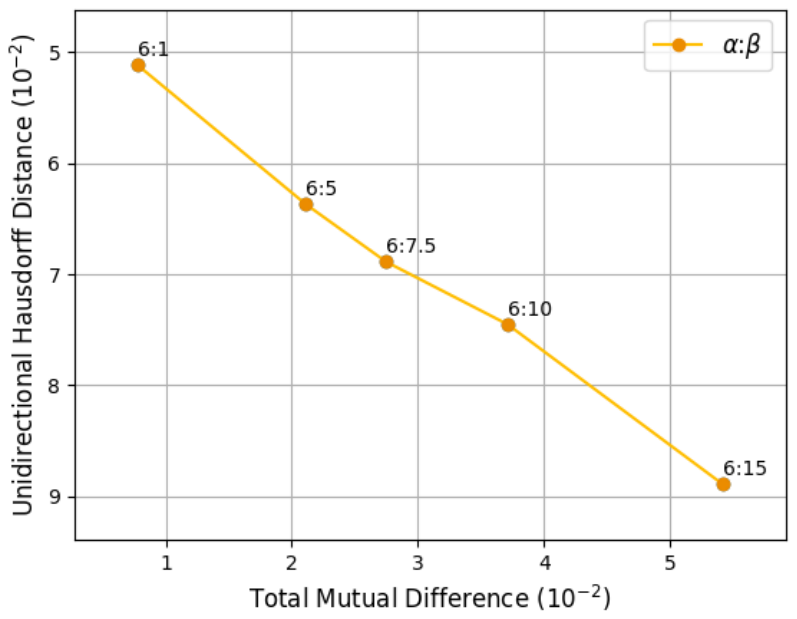}
    \end{minipage}
    \begin{minipage}[t]{0.45\textwidth}
		\centering
    	\includegraphics[width=0.9\textwidth]{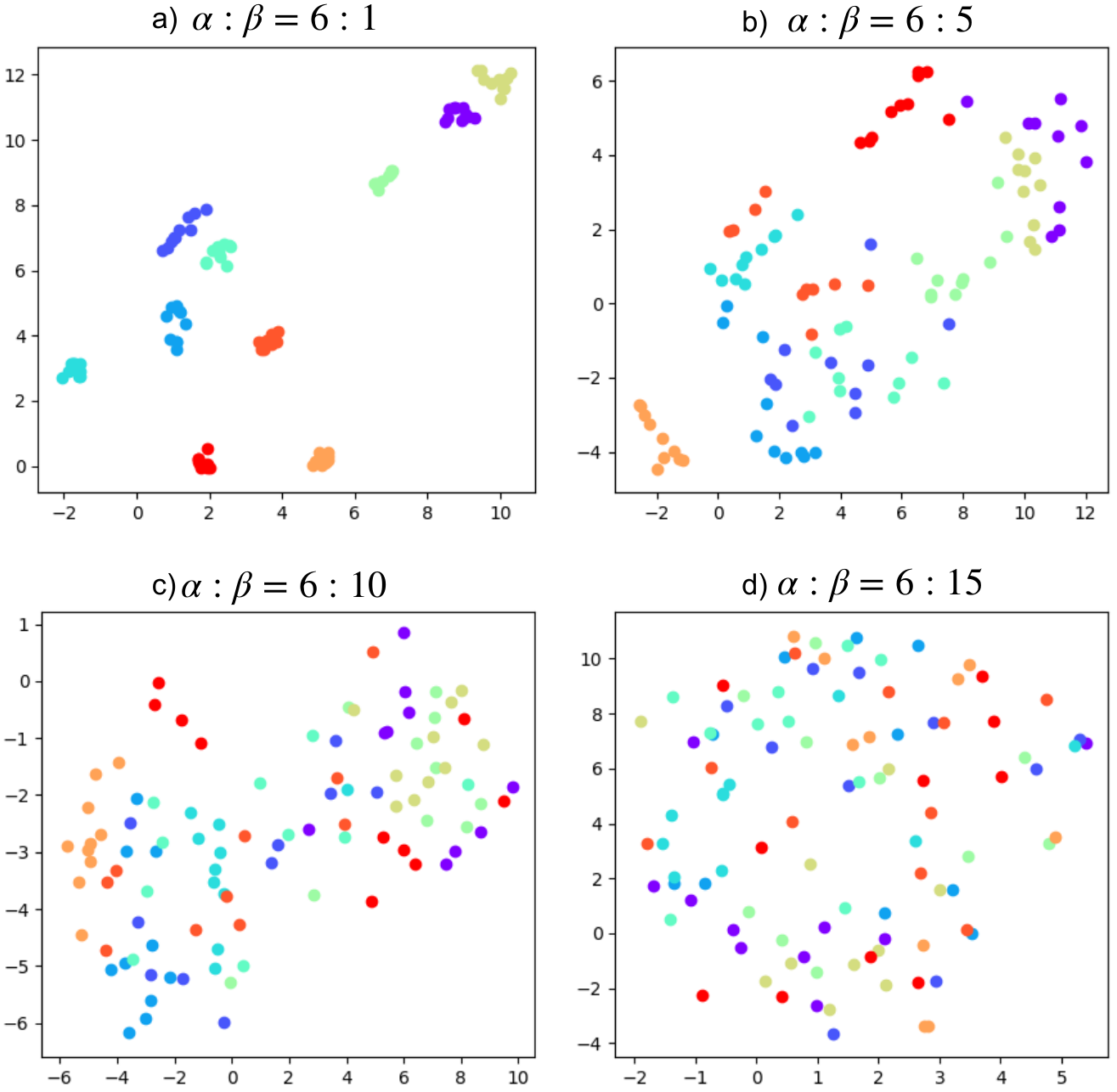}
    \end{minipage}
    \caption{Effect of trade-off parameters.
    Left: the completion fidelity (\texttt{UHD}) decreases and the completion diversity (\texttt{TMD}) increases as we set $\beta$ to be larger.
    Right: t-SNE visualization of completion latent vectors under different parameter settings. Dots with identical color indicates completion latent vectors resulted from the same partial shape. 
    a) setting a larger weight for $\LossRecon$ to encourage completion fidelity leads to mode collapse within the completion results of each partial input; 
    b)-d) setting larger weights for $\LossLatent$ to encourage completion diversity results in more modes.
    }
    \label{fig:weigt-effect}
\end{figure*}

\textbf{Effect of trade-off parameters.} 
Minimizing the $\LossGAN$, $\LossRecon$ and $\LossLatent$ in Eq.~\ref{eqn:obj_function} corresponds to the improvement of completion quality, fidelity, and diversity, respectively.
However, conflict exists when simultaneously optimizing these three terms, e.g., maximizing the completion fidelity to the partial input would potentially compromise the diversity presented in the completion results, as part of the completion point set is desired to be fixed. Hence, we conduct experiments and show in Fig.~\ref{fig:weigt-effect} to see how the diversity and fidelity vary with respect to the change of trade-off parameters $\alpha$ and $\beta$.

\begin{figure*}[h]
    \centering
        \begin{minipage}[t]{0.4\textwidth}
            \centering
            \includegraphics[width=0.8\textwidth]{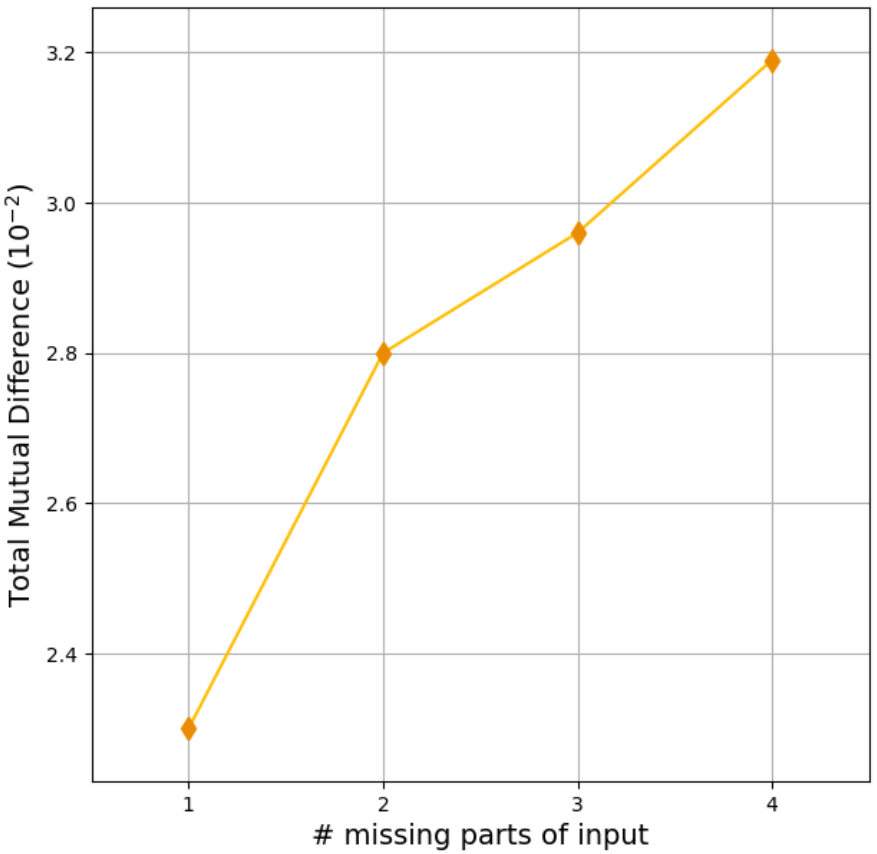}
        \end{minipage}%
        \begin{minipage}[t]{0.45\textwidth}
            \centering
            \includegraphics[width=0.9\textwidth]{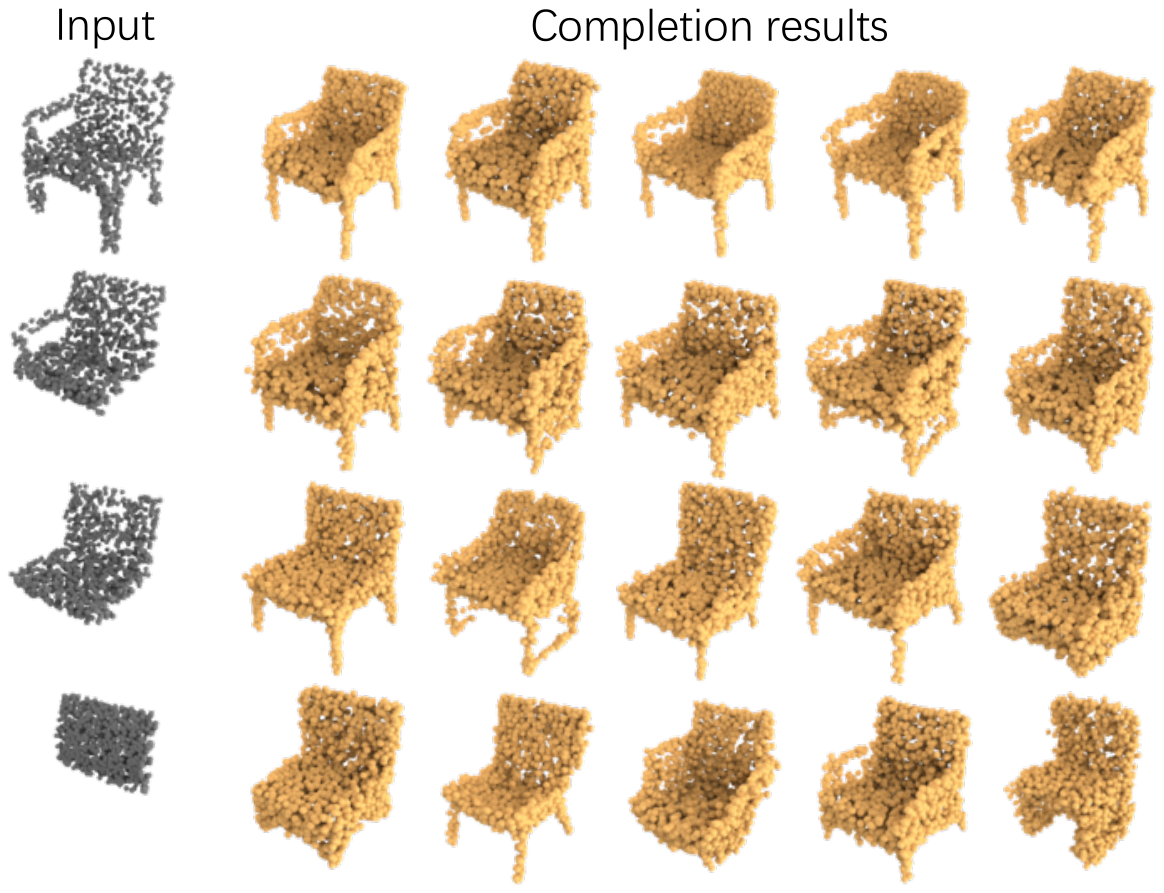}
        \end{minipage}%
    
        \caption{Effect of input incompleteness. 
        Completion results tend to show more variation as the input incompleteness increases.
        Left: the completion diversity (\texttt{TMD}) increases as the number of missing parts in the partial input rises.
        Right: we show an example of completion results for each input incompleteness. 
        }
        \label{fig:incompleteness-effect}
\end{figure*}

\textbf{Effect of input incompleteness.} 
The shape completion problem has more ambiguity when the input incompleteness increases. 
Thus, it is desired that the model can complete with more diverse results under increasing incompleteness of the input.  
To this end, we test our method on \texttt{PartNet} dataset, where the incompleteness can be controlled by the number of missing parts. Figure~\ref{fig:incompleteness-effect} shows how diversity in our completion results changes with respect to the number of missing parts in the partial input.


\section{Conclusion}
We present a point-based shape completion framework that can produce multiple completion results for the incomplete shape.
At the heart of our approach lies a generative network that completes the partial shape in a conditional generative modeling setting.
The generation of the completions is conditioned on a learned mode distribution that is explicitly distilled from complete shapes.
We extensively evaluate our method on several datasets containing different forms of incompleteness, demonstrating that our method consistently outperforms other alternative methods by producing completion results with both high diversity and quality for a single partial input.

The work provides a novel approach to indoor scene modeling suggestions and large-scale scene scan completions, rather than object-oriented completions.
While the completion diversity in our method is already demonstrated, the explicit multimodality encoding module is, nonetheless, suboptimal and could be potentially improved.   
Our method shares the same limitations as many of the counterparts: not producing shapes with fine-scale details and requiring the input to be canonically oriented.
A promising future direction would be to investigate the possibility of guiding the mode encoder network to concentrate on the missing regions for multimodality extraction, alleviating the compromise between the completion diversity and the faithfulness to the input.

\vspace{-5pt}
\section*{Acknowledgement}
We thank the anonymous reviewers for their valuable comments. This work was supported in part by National Key R\&D Program of China (2019YFF0302902) and NSFC (61902007).

%
%
\bibliographystyle{splncs04}
\bibliography{egbib}

\newpage

\begin{center}
\textbf{\large Supplementary Material}
\end{center}

\setcounter{section}{0}
 \renewcommand{\thesection}{\Alph{section}}
\newcommand{\ie}{\textit{i}.\textit{e}. }

\section{Overview}
This supplementary material contains:
\begin{itemize}
	\vspace{-5pt}
	\item implementation details of our network modules (Sec.~\ref{sec:network});
	\item details of the competing methods, which consists of the baseline methods and variants of our method (Sec.~\ref{sec:baseline}).
	\item the data processing details of the three datasets used in the evaluation (Sec.~\ref{sec:data}).
	\item details of the quantitative measures for the evaluation (Sec.~\ref{sec:metrics}). 
	\item more visual results of our multimodal shape completion (Sec.~\ref{sec:results-supp}).
\end{itemize}

\section{Details of network modules}
\label{sec:network}
Table \ref{table:network} shows the detailed architecture of our point set autoencoder($\Eae$, $\Dae$) and latent conditional GAN($G$ and $F$). 
Note that we fuse the latent code of the input partial shape and a Gaussian-sampled condition $\modez$ by direct channel-wise concatenation.

\begin{table}[h]
	\centering
	\footnotesize
	\begin{tabular}{|c|c|}
	\hline
	\multicolumn{2}{|c|}{\bf Pointnet Encoder $\Eae$} \\ \hline
Layer & Output Shape \\ \hline
Input point set & (3, K) \\ \hline
Conv1D+BN+ReLU & (64, K) \\ \hline
Conv1D+BN+ReLU & (128, K) \\ \hline
Conv1D+BN+ReLU & (128, K) \\ \hline
Conv1D+BN+ReLU & (256, K) \\ \hline
Conv1D+BN+ReLU & (128, K) \\ \hline
Global Max Pooling & (128,) \\ \hline

\multicolumn{2}{|c|}{\bf Decoder $\Dae$} \\ \hline
Layer & Output Shape \\ \hline
Latent code & (128, )\\ \hline
FC+ReLU & (256, ) \\ \hline
FC+ReLU & (256, ) \\ \hline
FC+ReLU & (2048$\times$3, ) \\ \hline
Reshape & (3, 2048) \\ \hline
	\end{tabular}
	\quad
	\begin{tabular}{|c|c|}
	\hline
	\multicolumn{2}{|c|}{\bf Generator  $G$} \\ \hline
Layer & Output Shape \\ \hline
Concat(Latent code, $z$)  & (128+64,) \\ \hline
FC+lReLU & (256,) \\ \hline
FC+lReLU & (512,) \\ \hline
FC & (128,) \\ \hline

	\multicolumn{2}{|c|}{\bf Discriminator $F$} \\ \hline
Layer & Output Shape \\ \hline
Latent code  & (128,) \\ \hline
FC+lReLU & (256,) \\ \hline
FC+lReLU & (512,) \\ \hline
FC & (1,) \\ \hline

	\end{tabular}
	\caption{Left: the architecture of our point set autoencoder. Right: the architecture of our latent conditional GAN. Conv1D: 1D convolution, BN: batch normalization, FC: fully connected layer, Concat: channel-wise concatenation. K is the number of points.}
	\label{table:network}
\end{table}

The variational autoencoder($\Evae$, $\Dvae$) shares the same architecture as the plain autoencoder ($\Eae$, $\Dae$), while having an extra FC layer at the bottleneck to squeeze the latent code to length $|\modez| = 64$ and enabling the re-parameterization trick.

\section{Details of competing methods}
\label{sec:baseline}
In this section, we describe in detail the design of baseline methods and variants of our method (namely \texttt{KNN-latent}, \texttt{ours-im-l2z} and \texttt{ours-im-pc2z}) in the comparison experimetns.

\begin{figure*}[t]
	\centering
	\includegraphics[width=\textwidth]{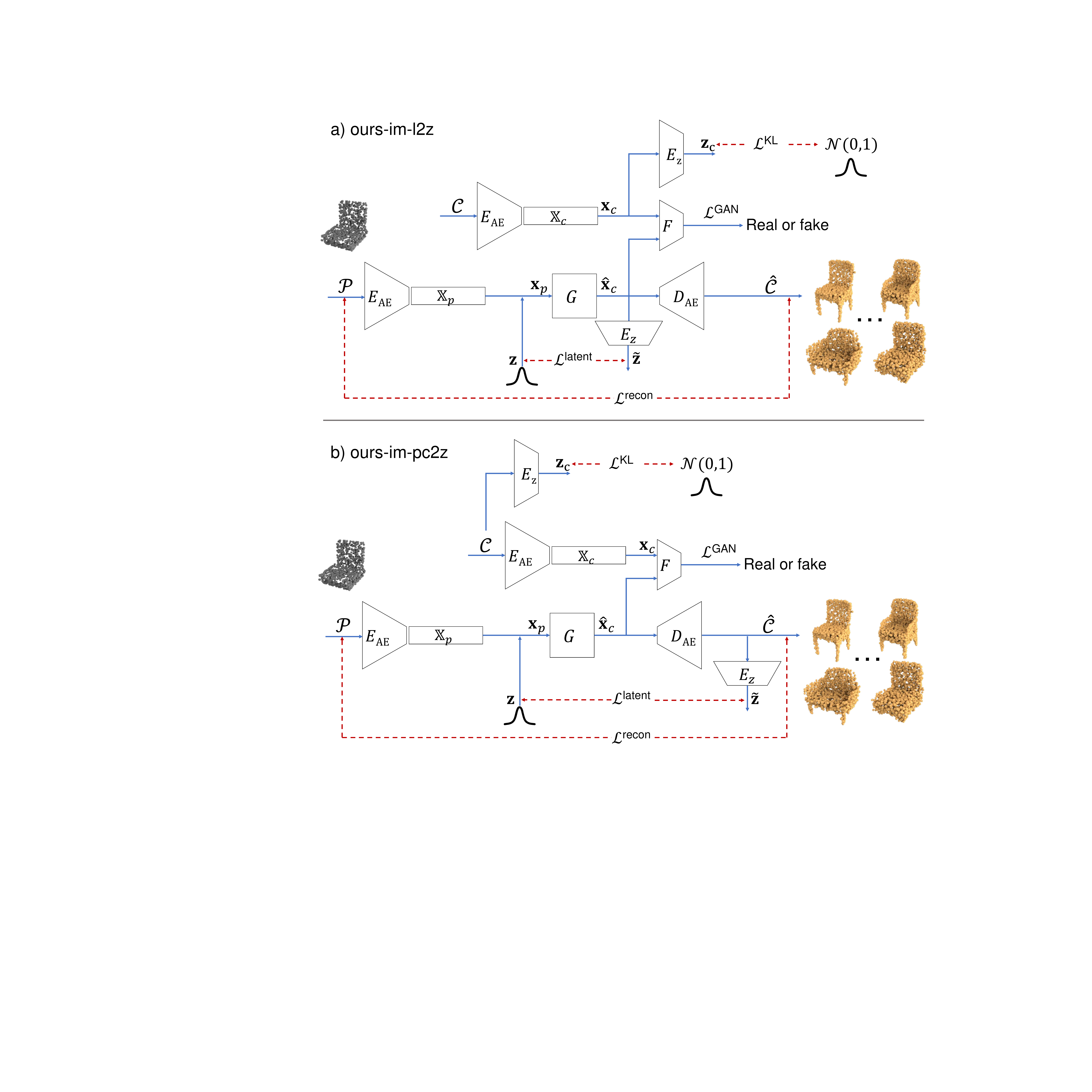}
	\caption{Illustration for the two variants of our method: \texttt{ours-im-l2z} (top) and \texttt{ours-im-pc2z} (bottom). }
	\label{fig:variants}
\end{figure*}

\texttt{KNN-latent}. Given an input partial shape, we encode it into the latent space formed by our point set encoder $\Eae$ and find its $k$-nearest neighbors based on cosine similarity.

\texttt{ours-im-l2z}. As a variant of our method, \texttt{ours-im-l2z} jointly trains the $\Ez$ to implicitly model the multimodality by mapping the complete latent code $\completecode$ into a low-dimensional space. Hence, the latent space reconstruction loss $\LossLatent_{{\generator, \Ez}}$ becomes:
$$\LossLatent_{{\generator, \Ez}} = \mathbb{E}_{\partialshape \sim p(\partialshape), \modez \sim p(\modez) } {\left[ \norm{\modez, \Ez(\generator(\Ecomp(\partialshape), \modez)) }_1 \right]}.$$
In addition to the loss terms of Eq. 6 in the main paper, to allow stochastic sampling at test time, an additional Kullback-Leibler (KL) loss on the $\modez$ space is introduced to force $\Ez(\completecode)$ to be close to a Gaussian distribution:
$$\LossKL_{\Ez}=\mathbb{E}_{\completecode \sim \completespace}[\mathcal{D}^{KL}(\Ez(\completecode)||\mathcal{N}(0,1))]$$
where $\mathcal{D}^{KL}$ stands for Kullback-Leibler divergence. Hence the full training objective function becomes:
\begin{equation}
	\argmin_{(\generator, \Ez)} \argmax_{\discriminator} \LossGAN_\discriminator + \LossGAN_\generator + \alpha \LossRecon_\generator + \beta \LossLatent_{{\generator, \Ez}} + \gamma \LossKL_{\Ez}
	\label{eqn:obj_function_l2z}
\end{equation}
The weight factors $\alpha$ and $\beta$ are set to $6.0$ and $7.5$, same as those of our main model, and $\gamma$ is set to $1.0$.

\texttt{ours-im-pc2z}. As stated in the main paper, \texttt{ours-im-pc2z} takes complete point clouds as input to implicitly encode the multimodality. 
The full training objective function is the same as that (Eq.~\ref{eqn:obj_function_l2z}) of \texttt{ours-im-l2z},
while the KL loss term changes to:

$$\LossKL_{\Ez}=\mathbb{E}_{\completecode \sim \completespace}[\mathcal{D}^{KL}(\Ez(\completeshape)||\mathcal{N}(0,1))],$$
The architectures for \texttt{ours-im-l2z} and \texttt{ours-im-pc2z} are shown in Fig.~\ref{fig:variants}.

\section{Details of data processing}
\label{sec:data}
In this section, we provide data processing details for our three datasets, especially the acquisition of the complete and partial point clouds in each dataset.

\texttt{PartNet}. Original PartNet dataset\cite{Mo_2019_CVPR} provides point clouds sampled from shape mesh surface, along with semantic label for each point. 
The provided point clouds of the complete shape serves directly as our complete shape data. 
To create the partial shape data, for a shape with $k$ parts, we randomly remove $j(1\le j \le k-1)$ parts by checking the semantic labels for all of its points.

\texttt{PartNet-Scan}. To resemble the scenario where the partial scan suffers from part-level incompleteness, both the complete and partial point sets in \texttt{PartNet\-Scan} are obtained from virtual scan.
Complete point sets are acquired by virtually scanning the complete shape mesh from $27$ uniform  views on the unit sphere. 
Partial point sets are acquired by virtually scanning the partial shape mesh from a single view that is randomly sampled. 
For each complete shape mesh in the original PartNet dataset\cite{Mo_2019_CVPR}, we create $4$ partial shape meshes using the same principle as in \texttt{PartNet}. 
And for each partial shape mesh, we run the single-view scan to get 4 partial point sets from different views.

\texttt{3D-EPN}. The provided point cloud representation directly serves as the partial shape data. The complete shape data comes from the virtual scan of ShapeNet\cite{shapenet2015} objects from 36 uniformly sampled views.

\section{Evaluation measures}
\label{sec:metrics}
\newcommand{\Tp}{\mathcal{T}_p}
\newcommand{\Tc}{\mathcal{T}_c}
\newcommand{\Gc}{\mathcal{G}_c}

Here we explain in detail the quantitative measures that we adopt for evaluation, \ie, Minimal Matching Distance \texttt{(MMD)}, Total Mutal Difference (\texttt{TMD}), and Unidirectional Haudorff Distance (\texttt{UHD}). 
Given a test set of partial shapes $\Tp$ and a test set of complete shapes $\Tc$. 
For each shape $p_i$ in $\Tp$, we generate $k$ completed shapes $c_{ij}$, $j=1...k$, resulting in a generated set $\Gc=\{c_{ij}\}$. We set $k=10$ in all our quantitative evaluations.

\texttt{MMD}~\cite{achlioptas2018learning}. For each shape $s_i$ in $\Tc$, we find its nearest neighbor $\mathbf{N}(s_i)$ in $\Gc$ by using Chamfer distance as the distance measure. \texttt{MMD} is then defined as 
$$\texttt{MMD} = \frac{1}{|\Tc|}\sum_{s_i\in \Tc} d^{\text{CD}}(s_i, \mathbf{N}(s_i)),$$
where $d^{\text{CD}}$ stands for Chamfer distance. In practice, we sample $2048$ points for each shape for calculation.

\texttt{TMD}. For each of the $k$ generated shapes $c_{ij}(1\le j \le k)$ from the same partial shape $p_i \in \Tp$, we calculate its average Chamfer distance to the other $k-1$ shapes and sum up the resulting $k$ distances. \texttt{TMD} is then defined as the average value over different input partial shapes in $\Tp$:
\begin{align*}
	\texttt{TMD} & = \frac{1}{|\Tp|}\sum_{i=1}^{|\Tp|}(\sum_{j=1}^{k} \frac{1}{k-1}\sum_{1\le l \le k,l\ne j} d^{\text{CD}}(c_{ij}, c_{il}))  \\
	& = \frac{1}{|\Tp|}\sum_{i=1}^{|\Tp|}(\frac{2}{k-1}\sum_{j=1}^k \sum_{l=j+1}^k d^{\text{CD}}(c_{ij})).
\end{align*}

\texttt{UMD}. We calculate the average unidirectional Hausdorff distance from the partial shape $p_i\in \Tp$ to each of its $k$ completed shapes $c_{ij}(1\le j \le k)$:
$$\texttt{UMD} = \frac{1}{|\Tp|}\sum_{i=1}^{|\Tp|}(\frac{1}{k}\sum_{j=1}^k d^{\text{HL}}(p_i, c_{ij})),$$
where $d^{\text{HL}}$ stands for unidirectional Hausdorff distance.

\newpage
\section{More results}
\label{sec:results-supp}
We show more results of our multimodal shape completion method.

\begin{figure*}[h]
    \centering
    \includegraphics[width=0.95\textwidth]{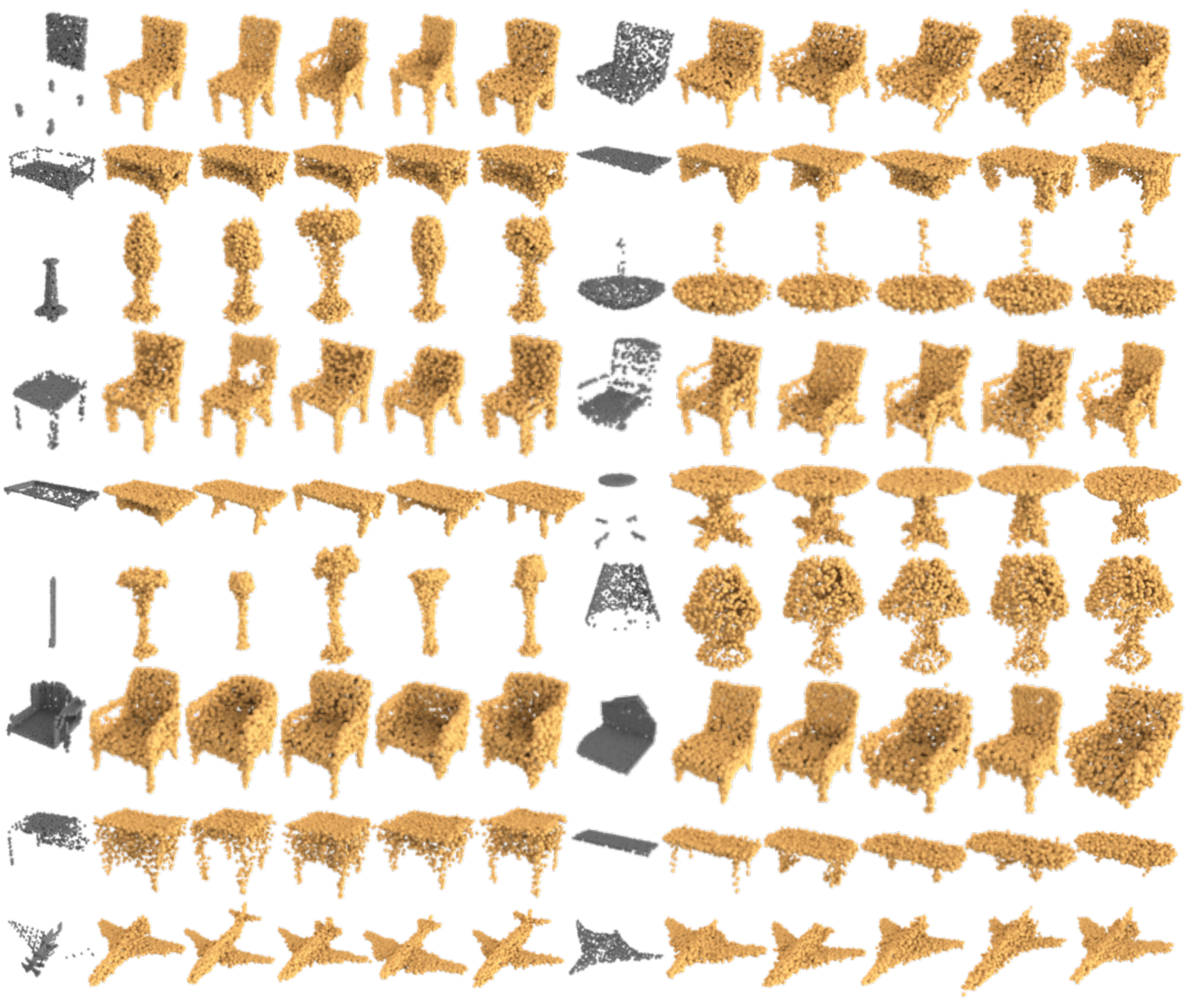}
    \caption{More results of our multimodal shape completion. The input partial shape is colored in grey, following by five different completions colored in yellow. From top to bottom: \texttt{PartNet} (rows 1-3), \texttt{PartNet-Scan} (rows 4-6), and \texttt{3D-EPN} (rows 7-9).}
    \label{fig:results-gallery-supp}
\end{figure*}

\end{document}